\pdfoutput=1

\documentclass[11pt]{article}

\usepackage{acl}

\usepackage{times}
\usepackage{latexsym}
\usepackage{multirow}
\usepackage{listings}
\usepackage[T1]{fontenc}

\usepackage[utf8]{inputenc}

\usepackage{microtype}

\usepackage{inconsolata}

\usepackage{comment}
\usepackage{xspace}
\usepackage{booktabs} 
\usepackage{subcaption}
\usepackage{graphicx}
\usepackage{amsmath}
\usepackage{courier}
\usepackage{xcolor}
\usepackage{amsthm}

\theoremstyle{definition}
\newtheorem{definition}{Definition}[section]

\definecolor{mediumgreen}{RGB}{60, 179, 113}
\lstdefinelanguage{Jinja2}{
  morekeywords={},
  sensitive=false,
  moredelim=[s][\color{blue}]{\{}{\}},
  moredelim=[s][\color{blue}]{\%}{\%},
  moredelim=[s][\color{mediumgreen}]{\{####}{####\}},
}

\newcommand{\TOCHECK}[1]{{\color{red} TOCHECK: #1}}

\newcommand{\suql}[0]{SUQL\xspace}
\newcommand{\summary}[0]{\textsc{summary}\xspace}
\newcommand{\classify}[0]{\textsc{classify}\xspace}
\newcommand{\answer}[0]{\textsc{answer}\xspace}
\newcommand{\textT}[0]{\texttt{text}\xspace}

\definecolor{palo_alto}{RGB}{23, 94, 84}
\newcommand{\hl}[1]{\textcolor{palo_alto}{\texttt{\textbf{#1}}}}
\newcommand{\textsql}[1]{\textrm{#1}}

\title{SUQL: Conversational Search over Structured and Unstructured Data \\with Large Language Models}

\author{Shicheng Liu \quad Jialiang Xu \quad Wesley Tjangnaka \\
{\bf Sina J. Semnani \quad Chen Jie Yu  \quad Monica S. Lam}
\\
Computer Science Department, Stanford University \\
Stanford, CA\\
\fontsize{11}{12}\selectfont\texttt{\{shicheng, xjl, wesleytj, sinaj, jyu01, lam\}@cs.stanford.edu}\\
}

\begin{document}
\maketitle
\begin{abstract}

While most conversational agents are grounded on either free-text or structured knowledge, many knowledge corpora consist of hybrid sources.
This paper presents the first conversational agent that supports the full generality of hybrid data access for large knowledge corpora, through a language we developed called SUQL (\textbf{S}tructured and \textbf{U}nstructured \textbf{Q}uery \textbf{L}anguage). Specifically, \suql extends SQL with free-text primitives (\summary and \answer), so information retrieval can be composed with structured data accesses arbitrarily in a formal, succinct, precise, and interpretable notation. With SUQL, we propose the first semantic parser, an LLM with in-context learning, that can handle hybrid data sources.

Our in-context learning-based approach, when applied to the HybridQA dataset, comes within 8.9\% exact match and 7.1\% F1 of the SOTA, which was trained on 62K data samples. 
More significantly, unlike previous approaches, our technique is applicable to large databases and free-text corpora. 

We introduce a dataset consisting of crowdsourced questions and conversations on Yelp, a large, real restaurant knowledge base with structured and unstructured data. 
We show that our few-shot conversational agent based on \suql finds an entity satisfying all user requirements 90.3\% of the time, compared to 63.4\% for a baseline based on linearization.\footnote{Code available at \url{https://github.com/stanford-oval/suql}}

\end{abstract}

\section{Introduction}

\begin{figure*}[ht]
  \centering
  \includegraphics[scale=0.62]{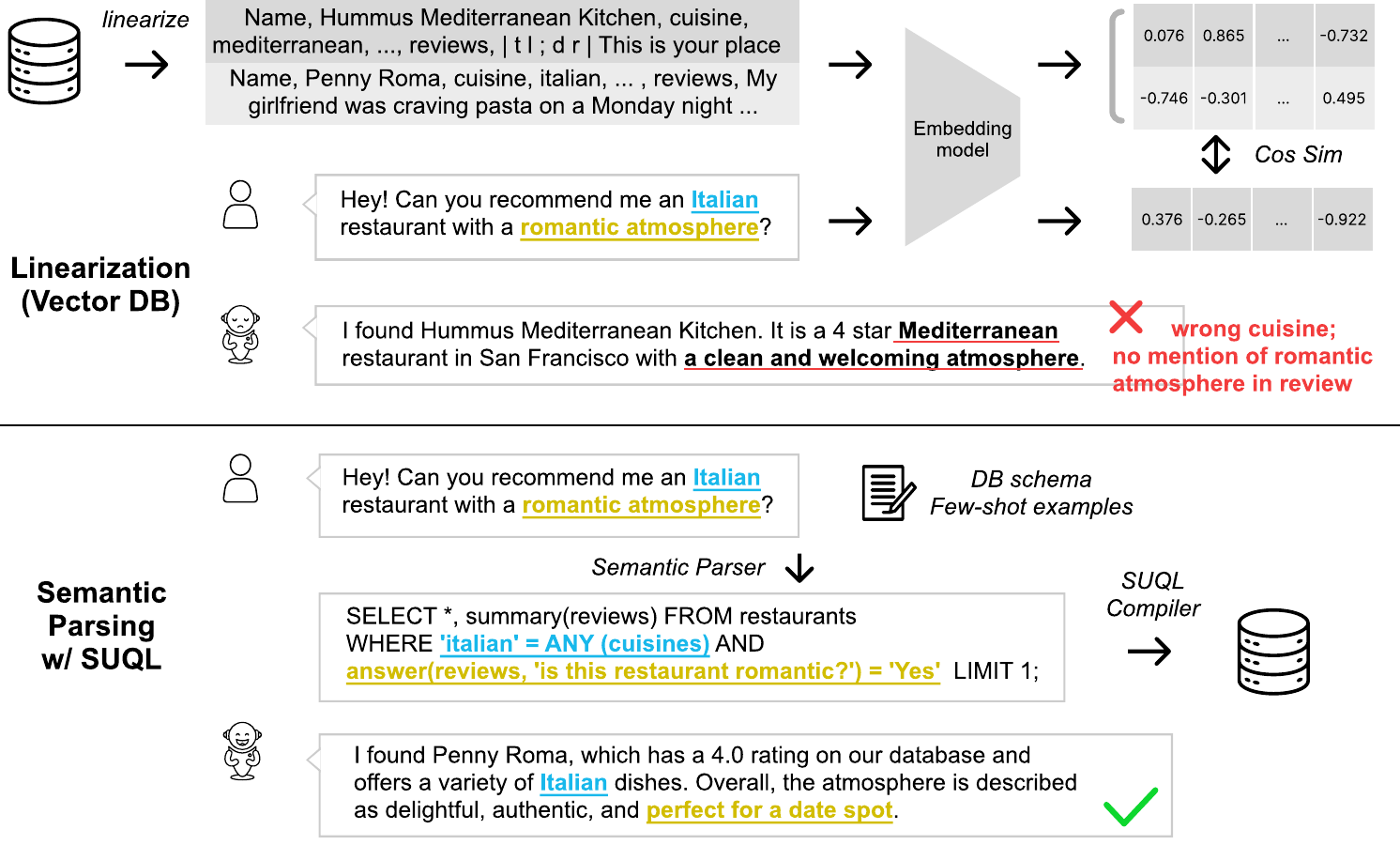}
  \caption{Comparison of traditional approach (linearization) with our approach (semantic parsing with SUQL).
  \\
  Top: In the linearization approach, database entries are linearized and converted to embedding vectors. At run-time, a user request is converted to an embedding vector, which is used to find the closest embedding from the stored vectors. The results are then supplied to LLM for response generation.
  \\
  Bottom: In our approach (semantic parsing with SUQL), a user utterance is parsed into formal SUQL by a few-shotted LLM, which is then executed by the SUQL compiler to fetch results from the database. The results are then supplied to LLM for response generation. \vspace{-0.1in}}
  \label{fig:figure1}
\end{figure*}
Large Language Models (LLMs) have shown exceptional performance on numerous downstream tasks. A range of recent works focus on improving their factuality by grounding responses in external resources including structured data~\citep{hu2022incontext, an2023skillbased, nan2023enhancing, poesia2022synchromesh, arora2023adapt, xu-etal-2020-autoqa, xu-etal-2023-fine} and free text~\citep{khattab2023demonstratesearchpredict, jiang2023active, semnani-etal-2023-wikichat, gao2023enabling}.


However, many data sources contain both structured data and free text: patient records, financial databases, and review websites, to name a few. 
Figure~\ref{fig:table1} in the appendix shows the running example of an application used in this paper. Each row in this table represents a unique restaurant, with information such as its name, type of cuisine, and rating as structured data. In addition, each row includes popular dishes and customer reviews in free text. To answer a question like ``Can you find me an Italian restaurant with a romantic atmosphere?'', an agent needs to combine the structured attribute cuisines and the free-text attribute reviews.

To handle the combination of structured and unstructured data, many previous chat systems use a \emph{classifier} to assign queries to one of its specialized modules that is designed to handle structured data, unstructured data, or chitchat~\cite{jin-etal-2021-assistance, chi-etal-2022-neural, zhao2023what}. Unfortunately, this approach is inadequate for questions that need both free-text and structured data.

Another popular approach is to convert, or \emph{linearize}, the structured data into free text~\citep{oguz-etal-2022-unik}, as shown in Figure~\ref{fig:figure1}. With this approach, we can no longer wield the power of SQL to query the database, and free text retrievers are not good at handling complex questions. 

The need of composing hybrid data source queries is highlighted by the HybridQA dataset, which collects many natural questions whose answers include information from both structured data and free text
\cite{chen-etal-2020-hybridqa}.  Previous attempts trying to ground question-answering systems on hybrid data~\citep{lei2023s,wu2023tacr, kumar-etal-2023-multi,hybridQA-lee-etal-2023-mafid} either work on only small datasets, or forego the expressiveness of structured data queries, or support limited compositions of structured and unstructured knowledge queries. 


This paper proposes an approach to grounding conversational agents in hybrid data sources that take full advantage of both structured data queries and free-text retrieval techniques. 

Our first contribution is to {\bf demonstrate empirically that in real-life conversations, it is natural for users to ask questions that span both structured and unstructured data}. Through crowdsourcing, we obtain questions that users ask and conversations that they have with a restaurant chatbot. Results show that more than 49\% of those questions require knowledge from both structured and unstructured knowledge.

To leverage the expressiveness and precision of formal query languages, {\bf we propose \suql, a precise, succinct, compositional, expressive, and executable formal language.} \suql augments SQL with several primitives for processing free text. At a high level, \suql combines an off-the-shelf retrieval model and LLMs (for unstructured data) with the SQL semantics and operators (for structured data).

{\bf We validate our approach using the HybridQA data set}. 
Experiments on HybridQA show that a few-shot, \suql-based QA system comes within 8.9\% exact match and 7.1\% F1 to the SOTA model trained on over 62K data samples.

{\bf We have developed a fully operational conversational agent with a few-shot LLM-based semantic parser with \suql}, shown in the bottom part of Figure \ref{fig:figure1}. We create a new single-turn user question data set and a conversational dataset on Yelp, a large, real knowledge corpus. Our chatbot using \suql finds an entity 
satisfying all user requirements 90.3\% of the time, compared to 63.4\% for a baseline based on linearization.

\section{Related Work}

{\bf Text-to-SQL Semantic Parsing.}
Text-to-SQL systems have been built for single-turn question answering tasks \citep{text-2-sql-guo2020content, text-2-sql-wang-etal-2020-rat, text-2-sql-scholak-etal-2021-duorat, text-2-sql-zhong2017seq2sql} as well as multi-turn, conversational tasks \citep{yu-etal-2019-cosql, yu-etal-2019-sparc, Wang2020TrackingIS, liu-etal-2022-augmenting-multi}. Recently, LLMs have shown promising results on the text-to-SQL semantic parsing problem via in-context learning \citep{brown-fewshot-2020}, with a range of work focusing on various prompting strategies \citep{hu2022incontext,  poesia2022synchromesh, an2023skillbased, nan2023enhancing, arora2023adapt, guo2023prompting, sun2023sqlprompt, zhang2023reactable}.


This line of work is only applicable to structured data sources without any free text. 
When it comes to free text, SQL is limited to basic pattern-matching on strings, hindering the application of text-to-SQL where deeper support is needed.

{\bf Specialized Modules Using a Classifier.}
One approach to building a conversational interface to hybrid sources is to classify each question and assign it to one of the specialized modules. For instance, Chirpy~\cite{chi-etal-2022-neural} implements different modules to handle different kinds of questions.
\citet{jin-etal-2021-assistance} and \citet{zhao2023what} implement a ``Turn Detection'' module, to determine whether a user turn involves unstructured data access or should be handled by APIs/DBs. However, real user questions naturally span across both structured and free text columns, which cannot be answered by systems built with separate modules.

{\bf Linearize structured data.}
Another popular approach is to turn structured data into a linear form, which can then be directly used by a language model. A common approach is to linearize raw tables row-by-row and feed linearized content into a Tabular Language Model (TaLM) \citep{herzig-etal-2020-tapas, yin-etal-2020-tabert, eisenschlos-etal-2021-mate, deng2022turl, iida-etal-2021-tabbie, sun2022iterative}. 

In particular, \citet{oguz-etal-2022-unik} linearizes Wikidata and Wikipedia tables, combines them with Wikipedia text, and applies a retrieval model to open-domain question-answering. However, this approach is inherently limited. Many queries are challenging to answer through free text alone, such as ``What are the number of deaths due to Covid in August and September 2020 in New York?''. These types of inquiries can be easily addressed using structured data, which supports comparisons and calculations across a big database. Moreover, linearization complicates the unification of different parts of the database. 

\section{Design and Rationale of SUQL}

\begin{table*}[htbp]
    \centering
    \resizebox{\textwidth}{!}{%
    \begin{tabular}{llll}

        \toprule
        Operator & Description & Example  \\
        \midrule
        \answer($t$ : \textT | \textT[], $q$ : \textT) & return the answer to & \answer\textsql{(reviews,}  \\
        $\to$ \textT &question $q$ on value $t$  & \textsql{"is this restaurant family-friendly?")} \\ \midrule
       \summary($t$ : \textT | \textT[]) $\to$ Text & return the summary of $t$ & \summary\textsql{(reviews)}  \\ 
        \bottomrule
    \end{tabular}
    }
    \caption{Free Text primitives in SUQL}
    \label{tab:free-text-operator}
\end{table*}

\label{sec:suql_design}
We present the design and rationale of the SUQL language in this section. The design objectives of the representation are {\em expressiveness}, {\em accuracy}, and {\em efficiency}.
\subsection{Design Rationale}
{\bf Expressiveness.} 
The design must be expressive, supporting the full generality of queries of hybrid knowledge corpus. It must handle arbitrary compositions of (1) relational operators in databases and (2) queries on free-text documents. Note that such a design automatically subsumes the {\em multi-hop retrieval} in NLP literature, where the results of a retrieved answer are used to retrieve another. 

Formal languages, such as SQL, have been proven to be {\em complete} with respect to relational algebra~\citep{Codd1972RelationalCO}. It can handle arbitrary compositions by virtue of its grammar rules, which for example, can be used to produce an unbounded number of nested subqueries.

Instead of linearization, which turns all structured data into text, we propose the opposite. SUQL is an extension of SQL with two NLP operators, \summary and \answer: \summary produces the summmary of a given text, and \answer returns the answer to a given text. These operators can be used anywhere with text values in the grammar, no different from numeric operators with numeric values. The advantage of this design is that SUQL is a succinct, formal representation that is complete with respect to relational algebra and NLP operations. 

{\bf Accuracy of Translation from Natural Language}.
LLMs have been shown to be capable of translating complex text in one natural language to another. They can translate complex sentences into SQL queries for albeit small databases with compound operations, such as the use of group-by, ranking, and subqueries.

We posit that SUQL will give LLMs a succinct notation to express complex queries involving hybrid data sources. Leveraging LLMs familiarity with SQL, we hypothesize that we can create a semantic parser for translating user queries in a conversation into SUQL queries with a LLM via in-context learning. 


{\bf Efficiency}. SUQL queries can be executed by the SQL compiler requiring no modifications, as the \summary and \answer primitives can be provided simply as user-defined functions. However, such an SQL compiler will perform very poorly. A naive implementation of these textual primitives would require retrieving and applying the NLP operation one record at a time, which is prohibitively expensive for large tables. Naive execution of the \answer function will not be effective.

Note that unlike previous methods such as retrieval-based semantic parsing where queries are constructed as results are retrieved~\citep{cao-etal-2022-program, gu-su-2022-arcaneqa}, SUQL expresses the query in its entirety. This makes it possible for us to develop an optimizing compiler, as described in Section~\ref{sec:optimize}. 

\begin{table*}[htp!]
\resizebox{\textwidth}{!}{%
\begin{tabular}{@{}cll@{}}
\toprule
\textbf{Question   Type} &
  \multicolumn{1}{c}{\textbf{Exemplar Question}} &
  \multicolumn{1}{c}{\textbf{SUQL query}} \\ \midrule
\begin{tabular}[c]{@{}c@{}}Type I\end{tabular} &
  Where was the XXXl Olympic held? &
  \begin{tabular}[c]{@{}l@{}} \textsql{\hl{SELECT answer}(``Event year Info'', `where is this event held?')} \\ \textsql{\hl{FROM} table \hl{WHERE} ``Name’' \hl{=} `XXXI’;}\end{tabular} \\ \midrule
\begin{tabular}[c]{@{}c@{}}Type II\end{tabular} &
  \begin{tabular}[c]{@{}l@{}} What was the name of the Olympic event held in Rio?\end{tabular} &
  \begin{tabular}[c]{@{}l@{}} \textsql{\hl{SELECT} ``Name’' \hl{FROM} table \hl{WHERE answer}(``Event year Info'',} \\ \textsql{`is this event held in Rio?') \hl{=} `Yes’;}\end{tabular} \\ \midrule
\begin{tabular}[c]{@{}c@{}}Type III\end{tabular} &
  When was the flag bearer of Rio Olympic born? &
  \begin{tabular}[c]{@{}l@{}}\textsql{\hl{SELECT answer}(``Flag Bearer Info'', `when is this person born?') \hl{FROM} table} \\ \textsql{\hl{WHERE answer}(``Event year Info'', `is this event held in Rio?') \hl{=} `Yes’;}\end{tabular} \\ \midrule
\begin{tabular}[c]{@{}c@{}}Type IV\end{tabular} &
  \begin{tabular}[c]{@{}l@{}}Which male bearer participated in Men’s 100kg\\  event in the Olympic game?\end{tabular} &
  \begin{tabular}[c]{@{}l@{}}\textsql{\hl{SELECT} ``Flag Bearer’' \hl{FROM} table \hl{WHERE} ``Gender'' \hl{=} `Male’ \hl{AND answer}(} \\ \textsql{``Flag Bearer Info'', `did this person participate in Men’s 100kg event?') \hl{=} `Yes';}\end{tabular} \\ \midrule
\begin{tabular}[c]{@{}c@{}}Type V\end{tabular} &
  \begin{tabular}[c]{@{}l@{}}For the 2012 and 2016 Olympic Event, when \\ was the younger flag bearer born?\end{tabular} &
  \begin{tabular}[c]{@{}l@{}}\textsql{\hl{SELECT MAX}(\hl{answer}(``Flag   Bearer Info'', `when is this person born?')::date)} \\ \textsql{\hl{FROM} table \hl{WHERE} ``Event year'' \hl{IN} (`2016', `2012');}\end{tabular} \\ \midrule
\begin{tabular}[c]{@{}c@{}}Type VI\end{tabular} &
  \begin{tabular}[c]{@{}l@{}}When did the youngest Burmese flag bearer \\ participate in the Olympic opening ceremony?\end{tabular} &
  \begin{tabular}[c]{@{}l@{}}\textsql{\hl{SELECT} ``Event year'' \hl{FROM} table \hl{ORDER BY answer}(``Flag Bearer Info'',} \\ \textsql{`when is this person born?')::date \hl{DESC LIMIT} 1;}\end{tabular} \\ \bottomrule
\end{tabular}%
}
\caption{The question types in HybridQA with exemplar questions (Figure 3 of \citet{chen-etal-2020-hybridqa}) translated to the corresponding SUQL queries. \vspace{-0.1in}}
\label{tab:hybridqa_suql}
\end{table*}
\subsection{Design of \suql}

We introduce two operations for text values in SQL. In this paper, we use \textT to represent any of the text types in SQL (CHAR, VARCHAR, TEXT, ...). 

We define \answer($t$,$q$) to return an answer to question $q$ on text input $t$.
For instance,
\vspace{-0.07in}
\begin{quote}
\answer(reviews, `is this restaurant family-friendly?') 
\end{quote}
\vspace{-0.07in}
will return yes if the reviews indicate that the restaurant is family-friendly, and no otherwise. The result is a text value that can be used anywhere it is allowed. 
For instance, 
\vspace{-0.07in}
\begin{quote}
\answer(reviews, `is this restaurant family-friendly?') \hl{=} `Yes' 
\end{quote}
\vspace{-0.07in}
can be used as a filter to select family-friendly restaurants.

\answer is a universal function that can be used to derive any information from a text value by supplying the right question. However, for convenience, we introduce \summary($t$) as syntactic sugar for
\vspace{-0.07in}
\begin{quote}
\answer($t$, `what is the summary of this document?').
\end{quote}
\vspace{-0.07in}
We posit it that the semantic parser can easily learn to use \summary.
The formal definitions of \answer and \summary are shown in Table~\ref{tab:free-text-operator}.


The \answer and \summary operations can be applied to any text arguments and their 
results can be used where a text value is expected, resulting in compositions of hybrid data accesses. Complex compositions of free text primitives and other SQL operators are highlighted by questions in the HybridQA dataset. In HybridQA, each cell in a column $C$ is potentially linked to some passages, which we store in a separate column called $C$\_Info. All questions from the dataset can be represented in \suql. We show 6 representative examples of how each type of question can be represented in \suql in Table \ref{tab:hybridqa_suql}.

\section{Conversational Agent}
\label{sec:conversational-agent}
Using SUQL as the formal representation, the architecture of a conversational agent with a hybrid knowledge corpus is relatively straightforward.


The Dialogue State Tracking problem~\cite{cheng-etal-2020-conversational, andreas-etal-2020-task, campagna-etal-2022-shot} for the SUQL-based conversational agent of a given schema $S$ is defined as follows. 
We define the dialogue history to consist of a sequence of utterances between the user and the agent, $A_1, U_1, \cdots, A_n, U_n$, where $A_i$ and $U_i$ denotes an agent utterance and user input at turn $i$, respectively. Each $U_i = (t_i, q_i)$ where $t_i$ is the natural text input, and $q_i$ is a \suql query for schema $S$ if $t_i$ carries a query.  Given schema $S$, ($A_i, U_i$) for all previous turns $1 \le i < n$ and the latest user utterance $t_n$, dialogue state tracking predicts $q_n$ if $t_n$ carries a query. 

The semantic parser for the dialogue state tracking consists of two stages, both implemented with an LLM using in-context learning. The first classifies if the knowledge corpus needs to be consulted. For user utterances like greetings or general questions, it skips the knowledge corpus access. 
If consulting is needed, the second stage predicts $q_n$. The prompt includes the schema definition and few-shot examples demonstrating \suql free-text primitives.

If the user utterance corresponds to a query, then the predicted \suql query is executed. Because the semantic parser may have translated the user query incorrectly, the agent is instructed via a prompt to explicitly state to the user what it searched, based on the predicted \suql at this turn (e.g., ``I searched for Italian restaurants with a romantic atmosphere.''). If the search returns a result, we ask the LLM to formulate the response based on the result; otherwise, we {\em explicitly} ask it to indicate that no results are found. The latter is important because LLMs tend to hallucinate whenever no answers to the user question are supplied. 


\section{An Optimizing \suql Compiler}
\label{sec:optimize}



Here, we describe the key optimizations we implemented in the SUQL compiler. 

\subsection{Search and Filter Optimization}

When \answer is used as a filter in the query, the naive implementation would require one LLM calling for every record in the database, which is infeasible. Similar to how database indexing is used to optimize queries, we use dense retrieval models to quickly identify the relevant records, instead of operating on them one-by-one. In addition, if only a few results are needed, it is unnecessary to evaluate the filter on all the records. 

First, our SUQL optimizing compiler identifies filters that use the \answer functions. It uses pre-computed embeddings from a dense retrieval model for similarity matching with the questions to identify top candidates. Note that the retrieved answers are relevant, but they may not satisfy the filtering constraints. We invoke the LLM on the entire clause to determine if the filter is successful.
For example, if the user asks for a restaurant where parking is easy, and a review happens to say ``the parking is hard''. The review may have the highest similarity score if no other reviews mention ``parking''. 
Thus, we need to apply the original filter \answer(reviews, `is parking easy?') = `Yes' using an LLM to ensure that the retrieved review indeed says that the parking is easy (Prompt \ref{prompt:suql-answer-filter}).

If multiple free text constraints are present, the \suql compiler uses an aggregated similarity score based on each constraint to retrieve results that most likely satisfy all constraints. Formally, the aggregated similarity score for each row $r$ is calculated as:
$$\sum_c \max_t \text{sim}(c, t)$$
where $c$ is a text constraint, $t$ is a text in row $r$, and $\text{sim}(\cdot, \cdot)$ denotes the similarity score between two texts.





\subsection{Enumerated Types}

Enumerated types (ENUM) are widely used in structured attributes to restrict the values of a text type to carry only one or more of a pre-defined set of permitted values. ENUM standardizes the values of the attributes so a filter based on the variable can be performed as a simple string match between the attribute values and permitted literals.  

The challenge is how to ensure that the semantic parser will map ENUM attribute values to a permitted one. For all ENUM type declarations with no greater than $N=10$ values, we include all the permitted values in the schema declarations supplied as a prompt to the LLM. The LLM is observed to be capable of automatically generating the ENUM values. 
For larger ENUM types, we do not include the permitted values, and the parser may generate an unexpected value.
For example, the user utterance ``Where can I find coffee'' is likely to be translated to the filter clause \textsql{`coffee' \hl{= ANY}(cuisines)}. However, the Yelp database only has `coffee \& tea' or  `cafe' cuisines, and not `coffee'. 

Our solution is to redefine the semantics of the \hl{=} operator for enumerated types. This is well known in the compiler literature as {\em overloading}. We first define the \classify function:  

\begin{definition}
\begin{align*}
\classify&(t:\textT, S:{v_1,\ldots, v_n})&\\
&=\{v_{i_1}, \ldots, v_{i_m}\},\forall v_{i_k} \in S \text{ similar to } t.
\end{align*}
\end{definition}
Here, we say two strings are similar if they have similar meanings. It is possible the value of interest is not included in the set of permitted values, in which case \classify returns $\emptyset$. We use a 0-shot LLM 
to implement \classify (Prompt \ref{prompt:enum-classifier}).

\begin{definition}
The equal operator \hl{=} is overloaded such that
\begin{align*}
t_1 = t_2 &\mbox{ iff } t_2 \in \classify(t_1, E), & \\
&\mbox{where } t_1:\textT \mbox{ and }t_2:\mbox{ENUM}(E)
\end{align*}
\end{definition}


For instance, given a clause \textsql{`coffee' \hl{= ANY}(cuisines)}, where \classify(`coffee', cuisines) = \{`coffee \& tea', `cafe'\}, then the clause will match any records whose cuisine attribute contains either `coffee \& tea' or `cafe'.

\subsection{Query Order Optimizations}
Since \answer and \summary involve LLM calls, it is important to minimize the execution of such functions. 

{\bf Predicate Ordering.}
As discussed above, \answer functions in filters are expensive, compared to other predicates. Thus, whenever possible, the \suql compiler would prioritize executing the other predicates so \answer is applied to fewer records.

Specifically, the \suql compiler converts \textsql{\hl{SELECT}} clauses with filter predicates into disjoint normal form (DNF), i.e., an OR of ANDs.  For each AND clause, it prioritizes filters not using the \answer function so \answer calls are applied only to the filtered records. 

{\bf Lazy Evaluation.}
Lazy evaluation, the concept of evaluating only when the result is needed, is a long-standing concept in programming languages \citep{10.1145/72551.72554}. The \suql compiler adopts this concept to minimize execution cost. Specifically, when a \textsql{\hl{LIMIT}} clause is present, it stops the evaluation once the required number of rows is filled.

\section{Experiments}
To evaluate SUQL, we perform two experiments. The first is on HybridQA, a popular academic question answering dataset as discussed above. Tables in HybridQA are small enough to be provided as input to a neural model. To perform a more comprehensive experiment on {\em conversations} with {\em large}, {\em real} data bases, we introduce a new benchmark based on the real restaurant data corpus from Yelp.com.

\subsection{HybridQA Experiment}

The HybridQA dataset consists of roughly 70K question-answering pairs aligned with 13,000 Wikipedia tables, whose entities are linked to multiple free-form corpora. Every question can be answered correctly only by referring to both the structured and unstructured data. To test out SUQL, we create the following system:

\begin{table*}[ht!]

\resizebox{\textwidth}{!}{%
\begin{tabular}{llllcll}
\toprule
\multirow{2}{*}{Method} & \multirow{2}{*}{\begin{tabular}{l}Model\end{tabular}}  & \multirow{2}{*}{Trained on (Size)}                 & \multicolumn{2}{c}{\textbf{Dev}} & \multicolumn{2}{c}{\textbf{Test}} \\
                  & & & EM              & F1             & EM              & F1              \\
\midrule
DocHopper~\citep{sun2022iterative} & \begin{tabular}{l}ETC\end{tabular} & \multirow{9}{*}{HybridQA (62k)} & 47.7            & 55.0           & 46.3            & 53.3            \\ \cmidrule(r){1-2} \cmidrule(l){4-7}
HYBRIDER~\citep{chen-etal-2020-hybridqa}      &\multirow{5}{*}{\begin{tabular}{l}BERT\end{tabular}} &   &  44.0            & 50.7           & 43.8            & 50.6            \\
MuGER$^2$~\citep{wang-etal-2022-muger2}        & & &  57.1            & 67.3           & 56.3            & 66.2            \\
Mate~\citep{eisenschlos-etal-2021-mate}  & & & 63.4            & 71.0           & 62.8            & 70.2            \\
DEHG~\citep{feng-etal-2022-multi}          & & & 65.2            & \textbf{76.3}  & 63.9            & 75.5            \\
MITQA~\citep{kumar-etal-2023-multi}           & & & 65.5            & 72.7           & 64.3            & 71.9            \\ \cmidrule(r){1-2} \cmidrule(l){4-7}
MAFiD~\citep{lee-etal-2023-mafid}            & \begin{tabular}{l}T5\end{tabular} & & 66.2            & 74.1           & 65.4            & 73.6            \\ \cmidrule(r){1-2} \cmidrule(l){4-7}
S$^3$HQA~\citep{lei-etal-2023-s3hqa} & \begin{tabular}{l}BERT/BART/DeBERTa\end{tabular} & & \textbf{68.4}   & 75.3           & \textbf{67.9}   & \textbf{75.5}   \\
\midrule
LLaMA2 (7B) ~\citep{zhang2023tablellama} & \multirow{2}{*}{\begin{tabular}{l} LLaMA2 (7B) \end{tabular}} & Zero-shot & 20.7 & - & - & - \\ \cmidrule(l){1-1} \cmidrule(l){3-7}
TableLlama~\citep{zhang2023tablellama} & & TableInstruct (2.6M)& 27.6 & - & - & - \\
\midrule
End-to-End QA w/ retriever ~\citep{shi2024exploring} & \multirow{3}{*}{\begin{tabular}{l} GPT-4 \end{tabular}}  & Zero-shot & 24.5\dag & 30.0\dag & - & - \\ \cmidrule(l){1-1} \cmidrule(l){3-7}
HPROPRO ~\citep{shi2024exploring} & & \multirow{2}{*}{Few-shot} & 48.0\dag & 54.6\dag & 48.7 & 57.7 \\
\suql (\textbf{Ours}) & & & 59.3 & 68.3 & 59.0 & 68.4 \\
\bottomrule
\end{tabular}
}
\caption{Performance of few-shot-based \suql and related work on the HybridQA dataset. \dag~denotes running on 200 sampled cases from the development set~\cite{shi2024exploring}.
}
\vspace{-0.1in} 
\label{tab:hybridqa_results}
\end{table*}

\begin{enumerate}
\setlength{\itemsep}{-3pt}
\setlength{\topsep}{-3pt}
\item 
Use LLM with in-context learning (with less than 10 examples) to parse natural language and a given database schema into a SUQL query (Prompt \ref{prompt:hybridqa-semantic-parser}).

\item
Execute the generated \suql to retrieve results from the database. If no results are returned, repeat this process by generating a different \suql query, with up to 2 tries (Prompt~\ref{prompt:hybridqa-no-result-recovery}).

\item
Use LLM to convert the retrieved database result to a succinct answer (Prompt \ref{prompt:hybridqa-format-extractor}) since the gold labels in HybridQA are short. Because the gold labels have only one entity, even though the full answer may include multiple entities, we just pick one out of the possibly many results returned by \suql. 
\end{enumerate}

\texttt{GPT-4-1106-preview} is used in all steps, except that \texttt{GPT-3.5-turbo-0613} is used in Step 3.

Our in-context learning-based QA system achieves 59.3\% Exact Match and 68.3\% F1 on the development set of HybridQA and 59.0\% EM and 68.4\% F1 on the held-out test set, as shown in Table~\ref{tab:hybridqa_results}. Our method uses only 3 simple prompts, achieving within 8.9\% EM and 7.1\% F1 to the SOTA on the test set, which has been trained on the HybridQA training set with over 62K examples. 

Most significantly, unlike our approach, these models do not generalize beyond small tables. 
Techniques based on feeding the entire table into a Transformer (DocHopper, Mate, MITQA, DEHG, and MAFID) cannot be applied to large data corpora that exceed their input token limit. Neither can techniques based on retrieving entire columns (MuGER$^2$) and feeding into a reader model. 
The SOTA model S$^3$HQA separately retrieves rows in the table and passages. It then feeds the top results of each to the final reader. It needs to feed the whole column to the reader if the query involves sorting. In contrast, our approach has full compositional generality and can handle arbitrarily large datasets. We are the first to apply semantic parsing techniques to HybridQA since no prior formal representations could accurately capture hybrid queries.


Recently, \citet{zhang2023tablellama} applied LLaMA-based techniques to HybridQA. They fine-tuned LLaMA2 (7B) on their TableInstruct dataset with more than 2.6M samples and achieved only 27.6 EM on Hybrid QA. They also reported a baseline of LLaMA2 (7B) on HybridQA directly, which resulted in just 20.7 exact match. 

\citet{shi2024exploring} reported two experiments of using GPT-4 on the HybridQA dataset. (1) Their GPT-4 End-to-End QA w/ retriever uses zero-shot GPT-4 to directly answer questions based on table and text parts retrieved by the retriever from \citet{chen-etal-2020-hybridqa}. (2) HPROPRO w/ GPT-4 uses few-shot program-based prompts to iteratively generate and execute Python code with the help of GPT-4. 
On a 200-sample of the development set, their two systems achieved 24.5 and 48.0 EM and 30.0 and 54.6 F1, respectively. The HPROPRO system achieves 48.7 EM and 47.7 F1 on the test set. 
Our SUQL-based approach outperforms both system, outperforming HPROPRO by more than 10.0\% in both EM and F1 on the test set.

\citet{sui2023gpt4table} also experimented using in-context learning with GPT-4 on HybridQA. 
However, they only reported the result of 1,000 randomly sampled questions from the development set. For each question, they experiment with different formats (JSON, HTML, Markdown, etc.) of feeding the entire table and question to GPT-4. 
The prediction is considered accurate if it is a substring of the gold answer, and vice versa\footnote{Based on communication with one of the authors.}. Their best-reported result is GPT-4 with HTML format at 56.68\% with this metric. Using their metric on the \textit{full} development set, our \suql-based system achieves a score of 72.5\%. 

These results show the effectiveness of \suql on the hybrid question-answering task, compared to other ICL techniques.



{\bf Error Analysis}.
From analyzing 72 randomly sampled error cases, we found:
\begin{itemize}
\setlength{\itemsep}{-3pt}
\setlength{\topsep}{-3pt}
\item 37.5\% are due to format mismatches, e.g. ``Johnson City, Tenessee'' versus ``Johnson City''. Similar issues related to evaluating LLM-generated responses have been noted by \citet{kamalloo-etal-2023-evaluating}. 

\item 23.6\% are due to the gold label being either wrong or incomplete. Incomplete cases exist because only one gold answer is permitted in HybridQA, while in fact for some cases, multiple possible correct answers could be found. 

\item 
22.2\% are due to semantic parsing errors. 

\item 
11.1\% are due to errors from the \suql execution involving the LLM-based \answer function and ENUM classifier. 

\item 
the remaining 5.6\% are due to type-related conversion errors, since HybridQA tables do not have annotated types while \suql expects a typed schema. 
\end{itemize}
In summary, even though the EM of \suql is 59.3\% on the development set, only 38.8\% of the 72 non-EM cases are true errors. Thus, the true accuracy of \suql may reach 84.2\%.

\subsection{Conversational Agent on Restaurants}
To experiment with real-life datasets, we collect a total of 1828 restaurants from Yelp.com across 4 cities, alongside the top 20 reviews and top 20 popular dishes for each restaurant. The columns of our database are name, cuisines, price, rating, num\_reviews, address, phone\_number, opening\_hours, location, reviews, and popular\_dishes.

We use an off-the-shelf dense retriever model \cite{yu-etal-2022-coco} 
as the retriever in \suql. We use \texttt{gpt-3.5-turbo-0613} as the LLM for all systems in this section. 

\subsubsection{Collecting User Queries}
We solicit user queries via crowdsourcing on Prolific~\cite{Prolific}. We do not disclose to the workers what fields are available in the database so as to not bias their queries. We ask them to come up with 100 questions about restaurants. Separately, we also ask crowd workers to interact with our conversational agent (described in Section~\ref{sec:conversational-agent}) and collect 96 turns across 20 conversations.

\begin{table}[htb]
    \small
    \centering
    \begin{tabular}{ccc}

        \toprule
         & Single-turn & Conversation \\
        \midrule
       Structured-only &45 & 37 \\
       Combination &55 & 25 \\
        \midrule
       Total  & 100 & 62   \\
        \bottomrule
    \end{tabular}
    \caption{Statistics on whether a search question requires only structured data or a combination.}
    \label{tab:test-stats}
\end{table}
\begin{table}[ht]
    \small
    \centering
    \begin{tabular}{ccc}
        \toprule
         & Single-turn & Conversational \\
        \midrule
       Linearization @ 1 & 57.0 \% & 63.4 \%\\
       Linearization @ 3 & 49.7 \% & 61.9 \%\\\midrule
       \suql         & \textbf{93.8 \%} & \textbf{90.3 \%} \\
        \bottomrule
    \end{tabular}
    \caption{Turn accuracy measurement on linearized system versus \suql system.\vspace{-0.2in}
    }
    \label{tab:precision}
\end{table}
The setting of restaurants in real-life use cases requires a user to first specify a location, a structured column in the database. We annotate whether a user question only involves structured information or a combination with free text in Table \ref{tab:test-stats}. In single turns, all collected user queries involve searching for a restaurant. Out of the 96 dialogue turns, 62 involve searching for restaurants. In total, over 49\% of user queries require knowledge from both structured and unstructured columns.

\subsubsection{Turn Accuracy}

We experiment with the linearization technique proposed by ~\citet{oguz-etal-2022-unik} for relational tables, using again
the same dense retriever model~\cite{yu-etal-2022-coco}. Specifically, we concatenate cell values on the same row and separate them by commas. Based on the conversation history, these systems use a few-shot LLM to extract a succinct search query for the retrievers. 

For each user input, we manually inspect whether the restaurants retrieved by a system satisfy all criteria specified by the user and respond with  correct and relevant information. Concretely, given a user utterance $u$ and a list of returned restaurants $\mathcal{R} = \{r_1, r_2, \cdots, r_m\}$, we evaluate  whether each $r_i$ 
is a true positive or false positive. We calculate the turn accuracy as the number of true positives divided by the number of true and false positives for all the queries in the dataset.

For the \suql system, the queries are limited to return at most 3 results. The accuracy is 93.8\% for single-turn questions and 90.3\% for conversational queries, as shown in Table \ref{tab:precision}. 

We compare our results  with two linearization-based systems, where $m=1$ (``Linearization \@ 1'') and $m=3$ (``Linearization \@ 3'').  \suql improves the answer accuracy, by up to 36.8\% in single-turn settings and up to 26.9\% in conversations. This shows that the conversational agent with \suql can provide much more accurate results.


Our system returns no answers to 21 of the 100 user questions and 8 of the 62 queries in the conversations.  Manual inspection reveals that 7 out of the 21 and 2 out of the 8 truly have no answers. Thus, our system has a false negative rate of 14\% and 9\% for user questions and conversational turns, respectively. 

\subsubsection{User Feedback}
We solicit feedback from our crowdsource users after they talk to our restaurant chat-bot with three free-form questions shown in Figure~\ref{fig:prolific_questions_after_chat}. 
Overall, the feedback was positive: ``There's actually nothing I didn't like about this chatbot. I would honestly use this chatbot on a regular basis if it were available to the public'', ``I liked that the chatbot was fast in responses and it gave very detailed responses and I hardly had any questions about a restaurant after the option was given'',  and ``Shocked at how good the restaurant suggestions were. I even asked for something with better prices and got that too. Now I'm hungry. I asked to define a cuisine style and it was able to do that''.

Negative comments include: occasional slowness of the chatbot; ``it didn't provide any links or pictures''; ``It did not sound friendly and sometimes the responses were too long. Bullet point outputs would be much more helpful.''

\section{Conclusion}

We introduce \suql, the first formal query language for hybrid knowledge corpora, consisting of structured and unstructured data.
The key novelty of \suql is the incorporation of free-text primitives into a precise, succinct, expressive, and interpretable query language. 

Our in-context learning based approach when applied to the HybridQA dataset comes within 8.9\% exact match and 7.1\% F1 to the SOTA on the test set trained on 62K data samples. 
More significantly, unlike previous approaches, our technique is applicable to large databases and free-text corpora. 

Our experiment on the real Yelp knowledge base with crowdsourced questions and conversations shows that our in-context learning conversational agent based on \suql finds an entity satisfying all user requirements 90.3\% of the time, compared to 63.4\% for a baseline based on linearization. The empirical findings underscore \suql's applicability and its potential for future research directions such as domain-specific applications in biomedical, legal, and financial spheres.

\section*{Ethical Considerations}

LLMs and formal languages such as SQL have been used by an increasingly large population of technical developers as well as everyday users. We propose to combine them in the hope of bringing the best of both sides to create a expressive, accurate, and efficient language that facilitates conversational search over structured and unstructured data. We do not foresee this work to result in any form of harm or malicious misuse. 

\textbf{Data.} The data used in this work is an open-sourced research dataset (HybridQA) and a Yelp-based restaurant conversation dataset (Restaurant). During the curation process of the Restaurant dataset, we used a certified online research crowdsourcing platform Prolific to make sure that we respected worker's privacy and paid them at fair rates. Our procedure has been approved by an IRB from our institution.


\textbf{Compute.} The models used herein are existing pretrained retriever models and LLM API services provided by OpenAI. We did not additionally pre-train or finetune any compute-intensive models, therefore avoiding a significant carbon footprint in the experiments herein.

\textbf{License. } Our code will be released publicly and licensed under Apache License, Version 2.0. Our data will be made available to the community.

\section*{Limitations}

Being LLM-based, SUQL can be subject to vulnerabilities that are intrinsic to LLMs. These intrinsic weaknesses can negatively affect SUQL's effectiveness, posing limitations on the overall pipeline performance. We highlight two aspects of limitation in the current version of SUQL methodology.

\textbf{Performance Limitation.} In this work, the LLM's semantic understanding capability upper-bounds the semantic and syntactic correctness of parsed SUQL queries. The \answer and \summary functionalities in SUQL can also be affected by the underlying LLM, resulting in potentially erroneous filtering evaluation during the execution of the SUQL queries. 

\textbf{Reliability Limitation.} The applicability of the method can also be affected by the reliability of the underlying LLM. In our pipeline, the semantic parser may hallucinate database contents in a non-interpretable manner, even when explicitly instructed not to. Other caveats include non-deterministic behavior between LLM API calls and potential vulnerabilities against LLM-oriented adversarial attacks.

\section*{Acknowledgements}
This work is supported in part by the National Science Foundation, the Alfred P. Sloan Foundation, the Verdant Foundation, Microsoft Azure AI credit, KDDI, JPMorgan Chase, and the Stanford Human-Centered Artificial Intelligence (HAI) Institute. We thank Gui Dávid for his experiments on the restaurants dataset in the early stages of the project. We also thank members of the Stanford OVAL (Open Virtual Assistant Lab) and the ACL ARR reviewers for their valuable comments and suggestions.

\bibliography{anthology,custom}

\appendix

\section{Appendix}


\begin{figure*}[ht]
  \centering
  \includegraphics[scale=0.6]{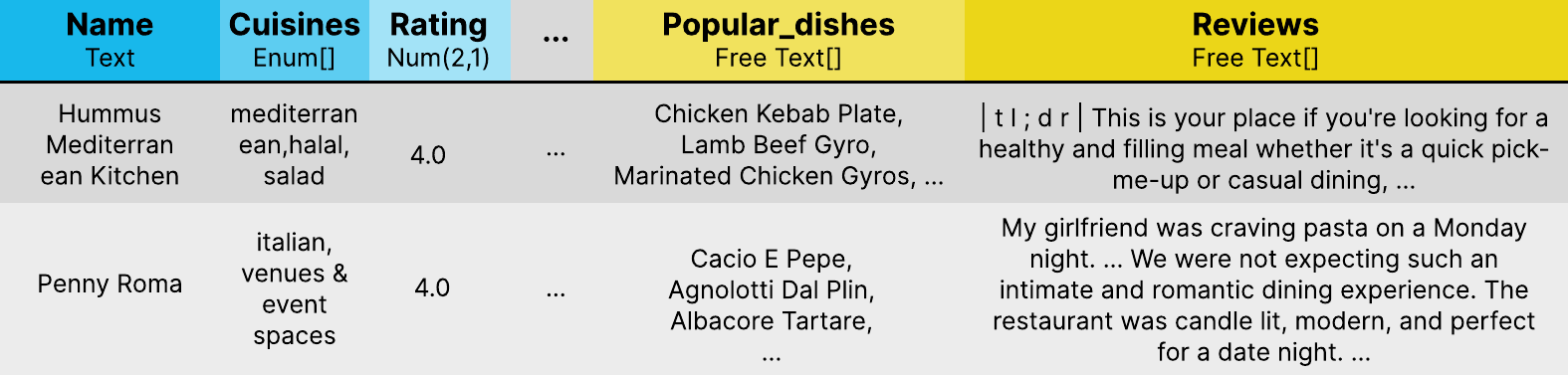}
  \caption{\texttt{restaurants} table with both structured and unstructured data.}
  \label{fig:table1}
\end{figure*}
\newpage
\subsection{Hyperparameters}
For all our experiments, we set a temperature of 0 in calls to OpenAI's LLMs, and we directly use the retriever provided by \citet{yu-etal-2022-coco}, with the default parameters.

\subsection{Prompts in our experiments}

We provide the prompts mentioned in this paper. The syntax used is the Jinja2 template language, which supports Python-like loops (\texttt{\{\% for \%\}\{\% endfor \%\}}), conditions (\texttt{\{\% if \%\}\{\% endif \%\}}), variables (\texttt{\{\{ var \}\}}) and comments (\texttt{\{\# \#\}}).

\begin{table*}
\begin{lstlisting}[basicstyle=\ttfamily\tiny]
In a database, the {{ field_name }} field has the following set of options, separated by new lines. "{{ predicted_field_value }}" is not one of the possible choices. You need to classify "{{ predicted_field_value }}" into one or more of the values below:

{% for choice in field_value_choices %}
{{ choice }}
{% endfor %}

You can only select from the above choices. Your response should be a list of comma separated index numbers.
Your answer: 
\end{lstlisting}
\caption{ENUM classifier prompt used in \suql compiler. This is a zero-shot prompt.}
\label{prompt:enum-classifier}
\end{table*}

\begin{table*}
\begin{lstlisting}[basicstyle=\ttfamily\tiny]
Answer a question based on the following text.{{ type_prompt }}

Question: {{ question }}. If there is no information, say "no info".

Documents:
{% for review in reviews %}
{{ review }}
{% endfor %}

Provide a concise answer in a few words:
\end{lstlisting}
\caption{The \answer function prompt used in \suql compiler. This is a zero-shot prompt.}
\label{prompt:suql-answer}
\end{table*}

\begin{table*}
\begin{lstlisting}[basicstyle=\ttfamily\tiny]
`answer(document, query)` takes in a document and a query. It asks `query` on `document` and outputs the answer.

Now, let's look at this use case. Your task is to determine whether the output is correct.

answer({{ field }}, "{{ query }}") {{ operator }} {{ value }}

{{ field }} = ["{{ document }}"]

Choose from one of the following choices:
- the output is correct.
- the output is incorrect.
\end{lstlisting}
\caption{The \answer function prompt as filter used in \suql compiler. This is a zero-shot prompt.}
\label{prompt:suql-answer-filter}
\end{table*}

\begin{table*}
\begin{lstlisting}[basicstyle=\ttfamily\tiny]
You are a semantic parser. Generate a query for a database with given signature. Do not generate fields beyond the given fields.

1929_International_Cross_Country_Championships_0
CREATE TABLE validation_table_7 ("Rank" INT, "Athlete" TEXT, "Athlete_Info" TEXT[], "Nationality" TEXT, "Nationality_Info" TEXT[], "Time" TEXT);
User: What is the difference in time between Jos\'e Reliegos of Spain and the person born 5 September 1892 who competed at the 1928 Olympics ?
Target: SELECT a."Time"::INTERVAL - b."Time"::INTERVAL FROM "validation_table_7" a, "validation_table_7" b WHERE a."Athlete" = 'Jos\'e Reliegos' AND a."Nationality" = 'Spain' AND answer(b."Athlete_Info", 'is this athlete born 5 September 1892?') = 'Yes';
--
List_of_cities_in_Somalia_by_population_0
CREATE TABLE validation_table_8 ("Rank" INT, "City" TEXT, "City_Info" TEXT[], "Region" TEXT, "Region_Info" TEXT[], "Population" INT);
User: Which gulf is north of the Somalian's city with 550,000 residents ?
Target: SELECT answer("City_Info", 'Which gulf is north of this Somalian''s city ?') FROM "validation_table_8" WHERE "Population" = '550,000';
--
List_of_the_mothers_of_the_Ottoman_Sultans_0
CREATE TABLE validation_table_14 ("Name" TEXT, "Name_Info" TEXT[], "Titles" TEXT, "Titles_Info" TEXT[], "Maiden Name" TEXT, "Origin" TEXT, "Origin_Info" TEXT[], "Death" DATE, "Son ( s )" TEXT, "Son ( s )_Info" TEXT[]);
User: Who was the husband of the mother of Ottoman sultan Suleiman I ?
Target: SELECT answer("Name_Info", 'Who is her husband?') FROM "validation_table_14" WHERE "Son ( s )" = 'Suleiman I';
--
List_of_Mohun_Bagan_A.C._managers_0
CREATE TABLE validation_table_10 ("Name" TEXT, "Name_Info" TEXT[], "Nationality" TEXT, "Nationality_Info" TEXT[], "FROM" DATE, "TO" DATE);
User: What is the nationality of the manager who was born on 15 February 1968 ?
Target: SELECT "Nationality" FROM "validation_table_10" WHERE answer("Name_Info", 'is this manager born on 15 February 1968?') = 'Yes';
--
Grammy_Award_for_Best_Jazz_Vocal_Performance,_Male_0
CREATE TABLE "validation_table_2615" ("Year" INT, "Year_Info" TEXT[], "Performing artist ( s )" TEXT, "Performing artist ( s )_Info" TEXT[], "Work" TEXT, "Work_Info" TEXT[], "Nominees" TEXT, "Nominees_Info" TEXT[])
User: How many people performed on the most recent song to win ?
Target: SELECT answer("Work_Info", 'how many people performed on this song?') FROM "validation_table_2615" ORDER BY "Year" DESC LIMIT 1;
--
List_of_flag_bearers_for_Myanmar_at_the_Olympics_0
CREATE TABLE validation_table_67 ("Name" TEXT, "Event Year" INT, "Year_Info" TEXT[], "Season" TEXT, "Flag Bearer" TEXT, "Flag Bearer_Info" TEXT[]);
User: When did the youngest Burmese flag bearer participate in the Olympic opening ceremony?
Target: SELECT "Event Year" FROM validation_table_67 ORDER BY answer("Flag Bearer_Info", 'when is this person born?')::date DESC LIMIT 1;
--
List_of_museums_in_Atlanta_0
CREATE TABLE validation_table_3 ("Name" TEXT, "Name_Info" TEXT[], "Area" TEXT, "Area_Info" TEXT[], "Type" TEXT, "Summary" TEXT, "Summary_Info" TEXT[]);
User: What is that address of the museum located in a Victorian House in an area whose Architectural styles within the district include Craftsman Bungalow , Queen Anne , Stick style , Folk Victorian , Colonial Revival , American Foursquare and Neoclassical Revival ?
Target: SELECT answer("Name_Info", 'what is the address?') FROM "validation_table_19" WHERE answer("Area_Info", 'is this an area whose Architectural styles within the district include Craftsman Bungalow , Queen Anne , Stick style , Folk Victorian , Colonial Revival , American Foursquare and Neoclassical Revival ?') = 'Yes';
--
2007_in_Canadian_music_0
CREATE TABLE "validation_table_26" ("Rank" INT, "Artist" TEXT, "Artist_Info" TEXT[], "Album" TEXT, "Album_Info" TEXT[], "Peak position" INT, "Sales" INT, "Certification" TEXT)
User: How many purchases of albums by the musician with the record Call Me Irresponsible have occurred ?
Target: SELECT answer("Artist_Info", 'How many albums has this artist sold?') FROM "validation_table_26" WHERE answer("Album_Info", 'is this record Call Me Irresponsible?') = 'Yes';
--
List_of_Indian_state_flowers_0
CREATE TABLE "validation_table_74" ("State" TEXT, "State_Info" TEXT[], "Common name" TEXT, "Common name_Info" TEXT[], "Binomial name" TEXT, "Binomial name_Info" TEXT[])
User:What is the state flower of the smallest state by area ?
Target: SELECT "Common name" FROM "validation_table_74" WHERE answer("State_Info", 'is this the smallest state by area?') = 'Yes';
--
List_of_Turner_Prize_winners_and_nominees_0
CREATE TABLE "validation_table_78" ("Year" INT, "Winner" TEXT, "Winner_Info" TEXT[], "Format" TEXT, "Nominees" TEXT, "Nominees_Info" TEXT[], "Notes" TEXT, "Notes_Info" TEXT[])
User: In what year did the 1999 Turner Prize winner win the Academy Award for his film , 12 Years a Slave ?
Target: SELECT answer("Winner_Info", 'in what year did he win the Academy Award for his film, 12 Years a Slave?') FROM "validation_table_78" WHERE answer("Winner_Info", 'did he win the Academy Award for his film, 12 Years a Slave?') = 'Yes' AND "Year" = '1999';
--
{{ table_original_name }}
{{ create_cmd }}
User: {{ query }}
Target: 
\end{lstlisting}
\caption{HybridQA semantic parser prompt. This prompt contains 10 examples, each with a (1) short table description, (2) table schema shown as a CREATE command, (3) the input query, and (4) the target \suql}
\label{prompt:hybridqa-semantic-parser}
\end{table*}


\begin{table*}
\begin{lstlisting}[basicstyle=\ttfamily\tiny]
You are a SQL semantic parser. In a prior turn, you have predicted a SQL, which returned no results. Your job now is to generate a new SQL to try again.
In addition to the standard SQL syntax, you can make use of the `answer` function.

In general, you should try to RELAX constraints.

Table description: Doping_at_the_Olympic_Games_15
Schema: CREATE TABLE "validation_table_56" ("Name" TEXT, "Name_Info" TEXT[], "Country" TEXT, "Country_Info" TEXT[], "Sport" TEXT, "Sport_Info" TEXT[], "Banned substance" TEXT, "Banned substance_Info" TEXT[])
Question: What substance was the athlete born in Bugulma banned in 2002 for using ?
Previously-generated SQL: SELECT "Banned substance" FROM "validation_table_56" WHERE answer("Name_Info", 'is this athlete born in Bugulma?') = 'Yes' AND "Country_Info" @> ARRAY['2002'];
This SQL returned no result.
New SQL: SELECT "Banned substance" FROM "validation_table_56" WHERE answer("Name_Info", 'is this athlete born in Bugulma and banned in 2002?') = 'Yes';
--
Table description: Sweden_at_the_1932_Summer_Olympics_0
Schema: CREATE TABLE "validation_table_1" ("Medal" TEXT, "Name" TEXT, "Name_Info" TEXT[], "Sport" TEXT, "Sport_Info" TEXT[], "Event" TEXT, "Event_Info" TEXT[])
Question: What was the nickname of the gold medal winner in the men 's heavyweight greco-roman wrestling event of the 1932 Summer Olympics ?
Previously-generated SQL: SELECT answer("Name_Info", 'What was his nickname?') FROM "validation_table_1" WHERE "Medal" = 'Gold' AND "Event" = 'Men''s heavyweight Greco-Roman wrestling';
This SQL returned no result.
New SQL: SELECT answer("Name_Info", 'What was his nickname?') FROM "validation_table_1" WHERE "Medal" = 'Gold' AND "Event" = 'Men''s heavyweight' AND "Sport" = 'Greco-Roman wrestling';
--
Table description: 2011_Berlin_Marathon_0
Schema: CREATE TABLE "validation_table_4" ("Position" INT, "Athlete" TEXT, "Athlete_Info" TEXT[], "Nationality" TEXT, "Nationality_Info" TEXT[], "Time" TIME)
Question: What place was achieved by the person who finished the Berlin marathon in 2:13.32 in 2011 the first time he competed in a marathon ?
Previously-generated SQL: SELECT "Position" FROM "validation_table_4" WHERE "Time" = '2:13:32' AND answer("Athlete_Info", 'is this the first time this person competed in a marathon?') = 'Yes';
This SQL returned no result.
New SQL: SELECT "Position" FROM "validation_table_4" WHERE "Time" = '2:13:32';
--
Table description: List_of_Pi_Kappa_Alpha_brothers_5
Schema: CREATE TABLE "validation_table_37" ("Name" TEXT, "Name_Info" TEXT[], "Original chapter" TEXT, "Original chapter_Info" TEXT[], "Notability" TEXT, "Notability_Info" TEXT[])
Question: What year was the brother from Beta Omicron born ?
Previously-generated SQL: SELECT answer("Name_Info", 'what year was this brother born?') FROM "validation_table_37" WHERE "Original chapter" = 'Beta Omicron';
This SQL returned no result.
New SQL: SELECT answer("Name_Info", 'what year was this person born?') FROM "validation_table_37" WHERE "Original chapter" = 'Beta Omicron';
--
Table description: List_of_radio_stations_in_the_United_Kingdom_15
Schema: CREATE TABLE "validation_table_55" ("Name" TEXT, "Name_Info" TEXT[], "Licence area" TEXT, "Licence area_Info" TEXT[], "Analogue frequencies" FLOAT, "Notes" TEXT)"
Question: Which station broadcasts to a civil parish in north west Dorset sited on the River Yeo ?
Previously-generated SQL: SELECT "Name" FROM "validation_table_55" WHERE answer("Licence area_Info", 'does this station broadcast to a civil parish in north west Dorset sited on the River Yeo?') = 'Yes';
This SQL returned no result.
New SQL: SELECT "Name" FROM "validation_table_55" WHERE answer("Licence area_Info", 'is this a civil parish in north west Dorset sited on the River Yeo?') = 'Yes';
--
Table description: {{ description }}
Schema: {{ schema }}
Question: {{ question }}
Previously-generated SQL: {{ previous_sql }}
This SQL returned no result.
{% if second_previous_sql is not none %}
    You also generated: {{ second_previous_sql }}
    This SQL also returned no result.
{% endif %}
New SQL: 
\end{lstlisting}
\caption{HybridQA no result recovery prompt. This prompt contains 5 examples, each with a (1) short table description, (2) table schema shown as a CREATE command, (3) the input query, (4) a previously generated \suql which returned no results, and (5) the target \suql.}
\label{prompt:hybridqa-no-result-recovery}
\end{table*}


\begin{table*}
\begin{lstlisting}[basicstyle=\ttfamily\tiny]
You are a good answer extractor. Given a detailed answer to a question, you always extract an succinct answer. If no valid answers can be extracted, answer with "No Info". Do not generate answers that is not from the original detailed answer. The succinct answer should be the minimum span from the passage without modification. When copying the answer, do not use a half word.

Question: The driver who finished in position 4 in the 2004 United States Grand Prix was of what nationality ?
Detailed Answer: The driver, Jenson Alexander Lyons Button, is British.
Succinct Answer: British
--
Question: What is that address of the museum located in a Victorian House in an area whose Architectural styles within the district include Craftsman Bungalow , Queen Anne , Stick style , Folk Victorian , Colonial Revival , American Foursquare and Neoclassical Revival ?
Detailed Answer: The address of the Hammonds House Museum is 503 Peeples Street SW in the West End neighborhood of Atlanta, Georgia.
Succinct Answer: 503 Peeples Street SW
--
Question: What is the area of the national park whose terrain is extremely rugged and consists of sandstone peaks , narrow gorges , ravines and dense forests , in kilometers ?
Detailed Answer: 524 km
Succinct Answer: 524
--
Question: Which gulf is north of the Somalian city with 550,000 residents ?
Detailed Answer: The Gulf of Aden is north of this city.
Succinct Answer: Gulf of Aden
--
Question: Who was the husband of the mother of Ottoman sultan Suleiman I ?
Detailed Answer: Her husband is Selim I.
Succinct Answer: Selim I
--
Question: What are the symptoms of the titular syndrome in his 2009 movie ?
Detailed Answer: The text does not provide information on the symptoms of any syndrome.
Succinct Answer: No Info
--
Question: {{ query }}
Detailed Answer: {{ detailed_answer }}
Succinct Answer: 
\end{lstlisting}
\caption{HybridQA format extractor prompt. This prompt contains 6 examples, each with a (1) input query, (2) a detailed answer from \suql, and (3) a target succinct answer.}
\label{prompt:hybridqa-format-extractor}
\end{table*}

\begin{table*}
\begin{lstlisting}[basicstyle=\ttfamily\tiny]
You are a restaurant virtual assistant chatting with a user.
You can access a restaurant database to retrieve information about restaurants' cuisine, price (cheap, moderate, expensive, luxury), rating (1-5), num_reviews, location, popular_dishes, reviews, phone_number, and opening_hours.

=====
{# basic #}
You: Hi! How can I help you?
They: what is a good place to get brunch in Chicago?
[Check the database? Yes]
=====
{# ask for rating #}
They: show me a Chinese restaurant in upper east side, NY
You: I found the 4.5 star Calle Dao Chelsea. It is a Chinese and Asian Fusion restaurant that serves brunch and has a menu that reflects the history of Chinese migration to Cuba.
They: is it better than panda express?
[Check the database? Yes]
=====
{# follow up on phone number #}
They: what is a good seafood restaurant in Seattle?
You: I found The Pink Door, a 4.5 star seafood restaurant in Seattle. Reviews mention the grilled whole fish and clam pasta as seafood dishes served there.
They: Can you find their phone number?
[Check the database? Yes]
=====
{# have you heard? #}
They: have you heard of girl and the goat at Chicago?
[Check the database? Yes]
=====
{# do not check db when asking about general knowledge #}
They: I want a Spanish restaurant in Kansas City
You: I found the 4 star La Bodega. It has one of the best happy hours in the city and a cozy atmosphere that feels like you're in the countryside of Spain.
They: Do you speak Spanish?
[Check the database? No]
=====
{# do not check db when asking about general knowledge about dishes #}
They: can you find me a place that serves peking duck?
You: Sorry, I don't have that information.
They: ok, how about orange chicken instead?
You: Sorry, I don't have that information.
They: how about chicken?
You: I found the 4 star Roost & Roast. It is a take-out place with no indoor dining, but there are tables outside. The menu is simple and ordering is done via an iPad kiosk. They offer Thai-inspired dishes such as Hat Yai Fried Chicken, Pad Thai, and several other stir frys.
They: what is hat yai fried chicken?
[Check the database? No]
=====
{# show another one #}
They: hey! show me something in Washington D.C.
You: I found the 4 star Old Ebbitt Grill. It is a historic restaurant located in Washington D.C. that has been serving locals and tourists since 1856. The food is very good and the service is great, with a friendly and knowledgeable staff.
They: is there another one?
[Check the database? Yes]
=====
{# keep on checking database even if there are errors #}
They: have you heard of girl and the goat at Chicago?
You: Sorry. I don't have that information
They: have you heard of girl and the goat at Chicago?
[Check the database? Yes]
=====
You: Hi! How can I help you?
{% for dlg_turn in dlg %}
They: {{ dlg_turn.user_utterance }}
{% if dlg_turn.agent_utterance is not none %}
You: {{ dlg_turn.agent_utterance }}
{% endif %}
{% endfor %}
[Check the database?
\end{lstlisting}
\caption{Input Classifier prompt for the restaurant experiment. This prompt contains 8 examples.}
\label{prompt:input-classifier}
\end{table*}

\begin{table*}
\begin{lstlisting}[basicstyle=\ttfamily\tiny]
You are a semantic parser. Generate a query for a restaurant database with the following signature:

CREATE TABLE restaurants (
    name TEXT,
    cuisines TEXT[],
    price ENUM ('cheap', 'moderate', 'expensive', 'luxury'),
    rating NUMERIC(2,1),
    num_reviews NUMBER,
    address TEXT,
    popular_dishes FREE_TEXT,
    phone_number TEXT,
    reviews FREE_TEXT,
    opening_hours TEXT,
    location TEXT
);

Do not generate fields beyond the given fields. The `answer` function can be used on FREE_TEXT fields.

{# Basic example #}
User: Where is Burguer King?
Target: SELECT address, summary(reviews) FROM restaurants WHERE name ILIKE '%Burguer King%' LIMIT 1;
--
{# Basic example for cuisine, and follow up with restaurant names #}
User: what are some good-reviewed japanese restaurants in Kansas City?
Target: SELECT *, summary(reviews) FROM restaurants WHERE 'japanese' = ANY (cuisines) AND location = 'Kansas City' AND rating >= 4.0 LIMIT 3;
Agent: I found Sakura Sushi, Nami Ramen, and Kaze Teppanyaki.
User: What are their prices?
Target: SELECT name, price FROM restaurants WHERE (name ILIKE 'Sakura Sushi' OR name ILIKE 'Nami Ramen' OR name ILIKE 'Kaze Teppanyaki') AND location = 'Kansas City';
--
{# Usage of `answer` function on FREE TEXT field in both projection and filter #}
User: Show me a family-friendly restaurant that has burgers in D.C.
Target: SELECT *, summary(reviews), answer(reviews, 'is this restaurant family-friendly?') FROM restaurants WHERE answer(reviews, 'do you find this restaurant to be family-friendly?') = 'Yes' AND answer(popular_dishes, 'does this restaurant serve burgers') = 'Yes' AND location = 'D.C.' LIMIT 1;
Agent: I found Jason's steakhouse. Reviews mention kids love going there with their parents. It should be a great weekend dinner for you and your family.
User: What do the reviews say about the atmosphere in the restaurant?
Target: SELECT answer(reviews, 'What is the atmosphere?') FROM restaurants WHERE name ILIKE 'Jason''s steakhouse' AND location = 'D.C.' LIMIT 1;
--
{# Usage of `answer` function on popular_dishes #}
User: Find me a place with pasta in Nashville.
Target: SELECT *, summary(reviews) FROM restaurants WHERE answer(popular_dishes, 'does this restaurant serve pasta') = 'Yes' AND location = 'Nashville' LIMIT 1;
--
{# Usage of `answer` function on reviews #}
User: I love Chinese food. Find me a restaurant that doesn't have a long wait time.
Target: SELECT *, summary(reviews), answer(reviews, 'what is the wait time?') FROM restaurants WHERE 'chinese' = ANY (cuisines) AND answer(reviews, 'does this restaurant have short wait time?') = 'Yes' LIMIT 1;
--
{# Meaning of the word "popular", and follow up on fetching reviews #}
User: I want a popular restaurant in Napa, CA.
Target: SELECT *, summary(reviews) FROM restaurants WHERE rating >= 4.5 AND location = 'Napa, CA' ORDER BY num_reviews DESC LIMIT 1;
Agent: I found the 5.0 star Gui's vegan house. It has 2,654 reviews and reviews mention great atmosphere, quick and good service, and good food quality.
User: Give me the review that talk about good food quality.
Target: SELECT single_review FROM restaurants AS r, unnest(reviews) AS single_review WHERE name ILIKE 'Gui''s vegan house' AND answer(single_review, 'does this review mention good food quality?') = 'Yes' AND r.location = 'Napa, CA' LIMIT 1;
--
{# Usage of `answer` function on reviews #}
User: Which restaurants have a happy hour in Bakersfield?
Target: SELECT *, summary(reviews), answer(reviews, 'what is the happy hour here?') FROM restaurants WHERE location = 'Bakersfield' AND answer(reviews, 'does this restaurant have a happy hour?') = 'Yes' LIMIT 1;
--
{# Usage of `answer` function on reviews #}
User: i'm hungry, what should i have for lunch? I am looking for salmon in Chicago.
Target: SELECT *, summary(reviews) FROM restaurants WHERE answer(popular_dishes, 'does this restaurant serve salmon?') = 'Yes' AND location = 'Chicago' LIMIT 1;
Agent: I found the 4.5 star Daigo. It is a family-owned business that serves traditional Japanese cuisine.
User: Show me something else.
Target: SELECT *, summary(reviews) FROM restaurants WHERE NOT(name = 'Daigo') AND answer(popular_dishes, 'does this restaurant serve salmon?') = 'Yes' AND location = 'Chicago' LIMIT 1;
--
{% for dlg_turn in dlg[:-1] %}
{% if dlg_turn.genie_utterance is not none %}
User: {{ dlg_turn.user_utterance }}
Target: {{ dlg_turn.user_target }}
Agent: {{ dlg_turn.agent_utterance }}
{% endif %}
{% endfor %}
User: {{ query }}
Target: 
\end{lstlisting}
\caption{The semantic parser prompt for the restaurant experiment. This prompt contains 8 examples.}
\label{prompt:restaurants-semantic-parser}
\end{table*}

\subsection{Our crowdsourcing process on Prolific}

\begin{figure*}[ht]
  \centering
  \includegraphics[scale=0.6]{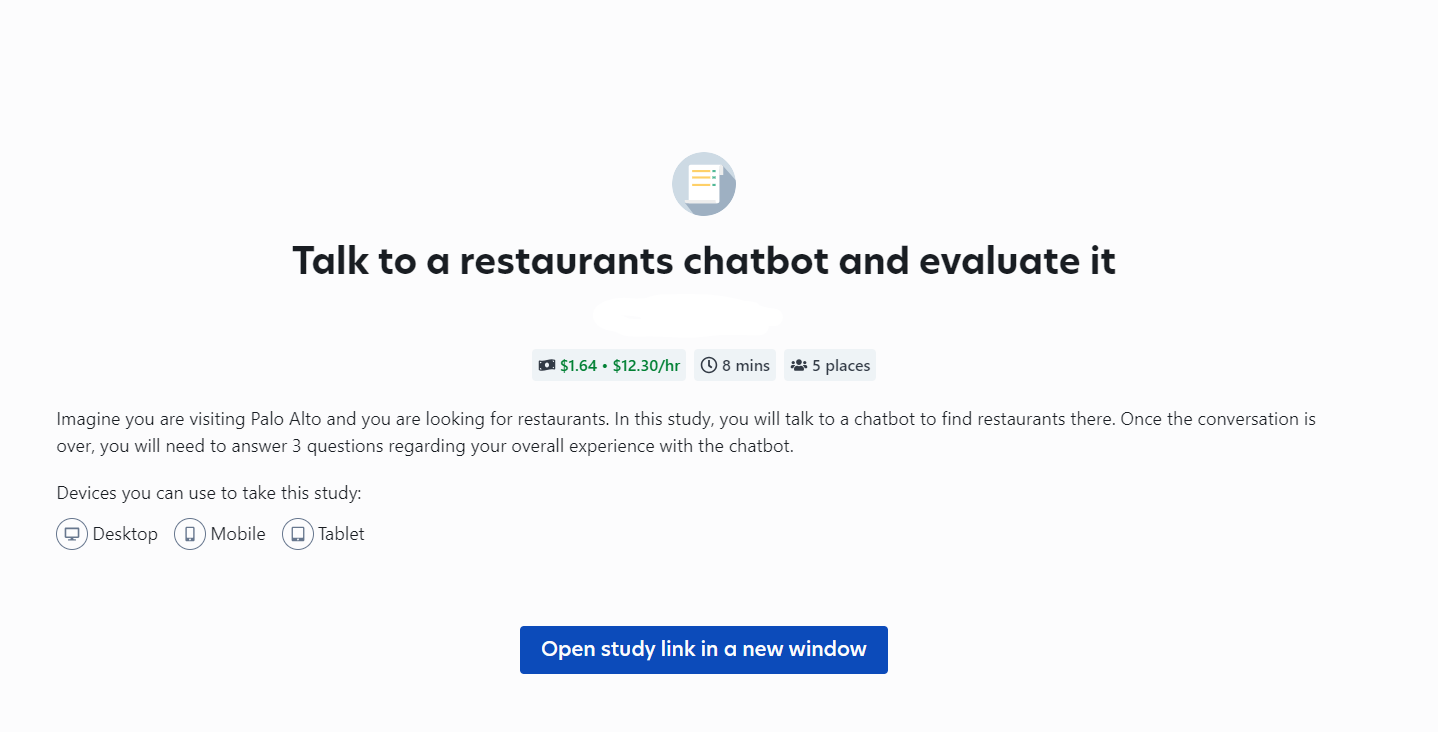}
  \caption{The crowdsourcing interface that our user sees}
  \label{fig:prolific_interface}
\end{figure*}

\begin{figure*}[ht]
  \centering
  \includegraphics[scale=0.6]{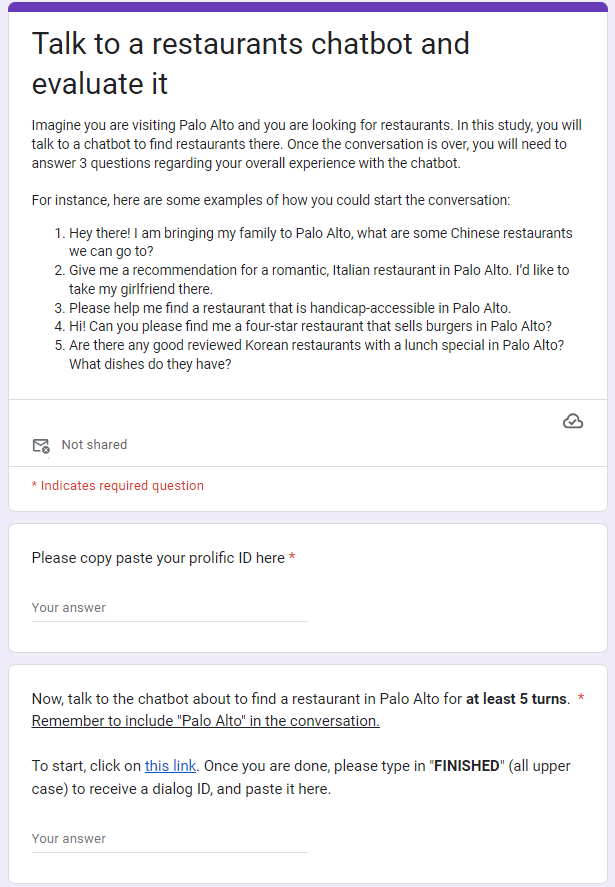}
  \caption{The prompts we give crowdsourcing workers before they start conversing with our chatbot.}
  \label{fig:prolific_first_page}
\end{figure*}

\begin{figure*}[ht]
  \centering
  \includegraphics[scale=0.6]{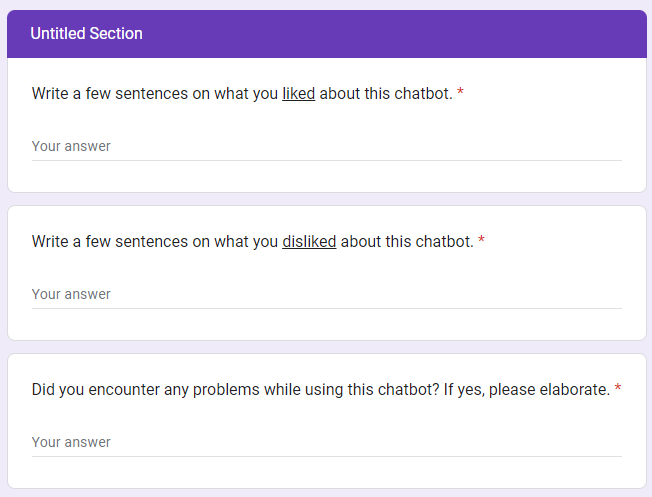}
  \caption{The questions crowdsourcing workers are asked after they finish talking to the chatbot.}
  \label{fig:prolific_questions_after_chat}
\end{figure*}

We utilize Prolific \citep{Prolific} to curate our Restaurant dataset. The crowdsourcing interface is presented in Figure~\ref{fig:prolific_interface}, after starting the crowdsourcing task, the crowdsourcing workers will be prompted with questions shown in Figure~\ref{fig:prolific_first_page}. After they finish conversing with the chatbot, they will be shown three questions shown in Figure~\ref{fig:prolific_questions_after_chat}. 

Among the 50 crowdsourcing workers who consented to reveal their demographic information, 33 are female and 17 are male. All 50 crowdsourcing workers reside in the United States. We paid the crowdsourcing workers 12.30 USD per hour. The average expected duration is 8 minutes. The pay rate is higher than the federal minimum wage in the United States, which is 7.25 USD per hour. Our crowdsourcing process asked for user consent in using their conversation with the chatbot for research purposes. No personal identifiable information was collected.

\clearpage

\end{document}